# Trustworthy synthetic data for campaign decision support: strategy simulation fidelity and the PolicySynth framework⋆

Tung Dang[a,*], The Hung Phung[b], Son Lam Nguyen[c] and Tu Nguyen[d,*]

[a] *Graduate School of Science, The University of Tokyo, 7-3-1 Hongo Bunkyo-ku, Tokyo, 113-0033, Japan*
[b] *Faculty of Human Resource Management, Trade Union University, 169 Tay Son, Kim Lien, Hanoi, Vietnam*
[c] *Institute for Business Management Development, Academy Of Finance, 58 Le Van Hien Street, Dong Ngac Ward, Hanoi, Vietnam*
[d] *Ministry of Industry and Trade, 23 Ngo Quyen, Hoan Kiem, Hanoi, Vietnam*

ABSTRACT

Decision support systems (DSS) increasingly run retention what-if analysis on synthetic customer populations, because privacy constraints preclude unrestricted use of real data. Such a system is trustworthy only if the synthetic data lead managers to the same decisions as the real data would; yet prevailing criteria certify distributional similarity, not decision alignment, so a synthetic population can match every marginal distribution while still steering a marketing team toward the wrong campaigns. We close this decision-alignment gap with three contributions: strategy simulation fidelity (SSF), a criterion measuring how often the synthetic population yields the same go/no-go campaign decision as the real population; PolicySynth, a DSS framework whose generator is conditioned on the production churn scorer to align decision-relevant structure; and a three-axis reporting standard of decision alignment, membership-inference resistance, and novel-record rate as the minimum deployment quality gate. On a telecommunications churn corpus and a banking acquisition corpus, PolicySynth attains a mean SSF of 0.923 and 0.960, with seed-to-seed variance roughly ten times tighter than CTGAN on telecommunications and 2.5 times on banking. This stability is the deployable property: go/no-go recommendations shift by at most ±1.2 percentage points between monthly retraining cycles, against ±11.5 for CTGAN, a reversed recommendation on one campaign in nine. A bootstrap baseline matches PolicySynth on SSF yet copies real records verbatim and fails membership inference, evidence that no single axis suffices. PolicySynth reliably supports directional go/no-go screening; its ROI estimates diverge from real outcomes by 70 to 78% and require the volume correction we document.

## 1. Introduction

Customer churn costs the telecommunications and subscription industries an estimated USD 35 billion a year [1, 2], and customer acquisition in retail banking is a comparably large, recurring marketing outlay. Every targeted campaign commits scarce budget against an uncertain response: a poorly targeted offer is spent on customers who would have acted anyway, and a poorly designed one fails to move those who are genuinely persuadable. Marketing decision-makers therefore want a decision support system (DSS) that can test a campaign design in simulation before any real budget is committed [3, 4, 5]. The natural data layer for such a system is a synthetic copy of the customer base: it supports unlimited what-if experiments while keeping personally identifiable records out of the analytics environment, as modern data-protection regimes increasingly require. Industry is already deploying such systems, but the DSS literature on synthetic-data-driven architectures is thin and its evaluation methodology has not kept pace [6, 7].

⋆

*Corresponding author
dangthanhtung91@g.ecc.u-tokyo.ac.jp (T. Dang); tunn@moit.gov.vn (T. Nguyen)
ORCID(s): 0000-0002-4974-9632 (T. Dang); 0009-0001-8682-5235 (T.H. Phung); 0009-0001-5302-5047 (S.L. Nguyen); 0009-0001-1781-2617 (T. Nguyen)
1

A synthetic-data layer helps a DSS only if its synthetic customers behave like real ones in the sense the decision depends on, and current practice does not measure behaviour in that sense. A synthetic population can match the real data on every standard statistical metric yet still lead a team to fund the wrong campaigns [8, 9], because a campaign decision turns on the joint relationship among churn probability, customer value, and contract structure, not on any single marginal. We call this the *decision-alignment gap*: the difference between distributional fidelity (the synthetic data resemble the real data statistically) and decision fidelity (the decisions made on the synthetic data agree with those made on the real data). No existing synthetic-data criterion measures it, so decision-makers have no principled way to compare candidate synthetic-data layers before putting one into production [10, 11, 12]. The gap is the evaluation-side counterpart of a principle already established on the training side, where decision-focused learning shows that models trained against a downstream decision loss outperform those trained against a generic predictive loss [13, 14, 15].

This paper makes three contributions to the DSS literature, forming one agenda, a criterion, a framework that satisfies it, and a reporting standard that operationalises both. First, strategy simulation fidelity (SSF) measures the share of strategies, within a parameterised family, on which the synthetic population yields the same go/no-go decision as the real one: a single number telling a deployment team how often the DSS's campaign recommendations will match those it would have made on real data. Second, because no existing generator is built to optimise SSF, we propose PolicySynth, whose generator is trained with a decision-conditioning loss derived from the production churn scorer; its operational value is stability, on the telecommunications corpus its recommendations shift by at most $\pm1.2$ percentage points between monthly retraining cycles, against $\pm11.5$ for the strongest competing architecture. Third, because decision alignment alone does not make a layer deployable, we establish a three-axis reporting standard, decision alignment (SSF), membership-inference resistance (MIA-AUC), and novel-record rate, as a deployment checklist any organisation can apply to a candidate synthetic-data tool before approval.

We evaluate this agenda on two real, structurally different customer bases, a telecommunications churn-retention corpus and a banking term-deposit-acquisition corpus, that differ in decision type, base rate, and value function, so that properties of the framework can be separated from properties of one decision problem. The stability and privacy results hold on both; the contribution of the decision-conditioning mechanism is corpus-dependent, load-bearing where the decision turns on a label-correlated joint and redundant where a single value magnitude dominates. We treat this not as a failure to generalise but as a scope condition the architecture makes predictable in advance.

Three research questions guide the investigation:

- **RQ1.** How should the quality of a synthetic-data layer in a DSS be evaluated when its purpose is to support managerial decisions rather than predictive modelling?
- **RQ2.** Which architectural choices in a synthetic-data generator are causally responsible for decision-alignment performance, and under what conditions on the decision problem does each matter?
- **RQ3.** What is the minimum reporting standard that lets decision-makers compare synthetic-data-driven DSS candidates before deployment?

The remainder of the paper is organised as follows. Section 2 reviews the three literatures that bear on evaluating synthetic-data-driven DSS. Section 3 formalises the decision-alignment gap, defines SSF, and proposes the three-axis standard. Section 4 presents the PolicySynth framework and its decision-conditioning architecture. Section 5 reports the empirical evaluation and deployment case study across both corpora. Sections 6 and 7 discuss implications and limitations and conclude.

## 2. Related work

Three literatures bear on evaluating synthetic-data-driven decision support: the DSS work on simulation-based what-if analysis, the tabular synthetic-data generation literature, and decision-oriented evaluation in adjacent fields. None has, on its own, produced a criterion that measures whether a synthetic-data layer supports the same managerial decisions as the real population.

### 2.1. Synthetic-data-driven decision support systems

Decision support systems have drawn on simulation for what-if analysis since the 1970s, generating the population on which strategies are tested from a process model, a discrete-event simulator, an agent-based model, or a system-dynamics representation [16, 17, 18]. A more recent strand replaces the rule-simulated population with one generated by statistical or machine-learning models trained on historical data [19, 20, 21]; we call these synthetic-data-driven DSS. Three properties distinguish them from their simulation-based predecessors: the data layer is trained on real production data, which adds realism but a privacy attack surface [22]; the generator is opaque relative to a rule-based simulator, so its decision-relevant behaviour is harder to audit [23]; and the layer must be refreshed as the population drifts, which imposes a stability requirement fixed-rule simulators never face. Existing DSS work has addressed how to integrate generative models into pipelines, not how to evaluate the learned data layer for managerial use [24, 25]. This is the first gap we close.

### 2.2. Tabular synthetic-data generation

The state of the art in tabular synthesis spans four method families: conditional GANs (CTGAN), variational autoencoders (TVAE), denoising diffusion models (TabDDPM), and autoregressive transformers (REaLTabFormer, GReaT) [26, 27, 28, 29]. Each is trained against a distributional objective, and the standard reporting toolbox is correspondingly distributional: per-column total-variation and Kolmogorov–Smirnov distances, correlation-matrix distance, train-on-synthetic-test-on-real (TSTR) AUC for utility, and membership-inference resistance for privacy [30, 31, 32]. None of these measures whether the decisions a team would make on the synthetic population match those it would make on the real one. The closest, TSTR, evaluates a downstream model rather than a distribution, but it scores predictive performance, not decision agreement, and the two diverge among otherwise competitive generators. This is the second gap.

### 2.3. Decision-oriented evaluation in adjacent fields

The principle that an artefact should be judged by its consequences for downstream decisions, not its intrinsic statistics, has been developed in three adjacent fields: off-policy evaluation in reinforcement learning [33], counterfactual evaluation in causal inference [34], and, most directly, decision-focused (predict-then-optimise) learning, which trains predictive models against the downstream decision loss and shows that decision quality and predictive quality are distinct objectives [13, 15, 35, 36, 37]. That work establishes the distinction on the training side; we establish its counterpart on the evaluation side, for the synthetic-data layer of a DSS [38, 39, 40]. The methodological lineage is established; what is new is the formalisation of decision alignment as a measurable evaluation criterion for tabular synthesis, within a DSS context.

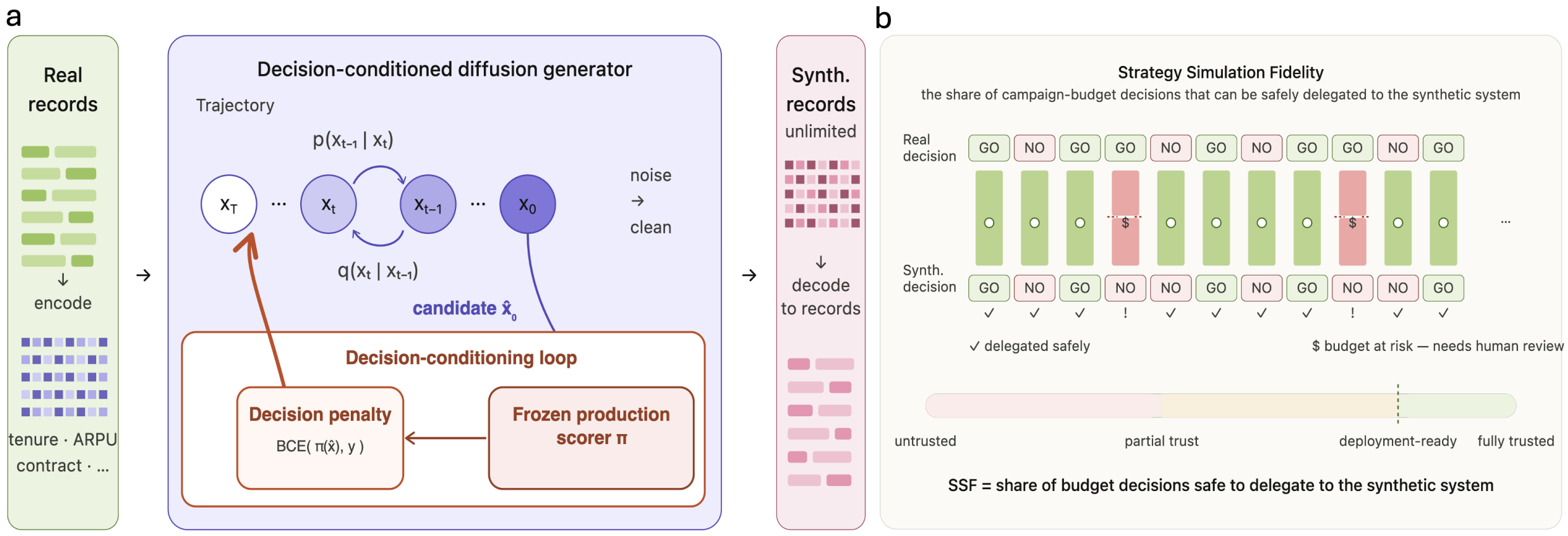

**Figure 1:** The PolicySynth framework and its evaluation construct. (a) Architecture. Encoded real customer records (tenure, ARPU, contract, ...) seed a denoising diffusion generator that maps a noisy draw $x_T$ to a clean synthetic record $x_0$ via the reverse process $p(x_{t-1}|x_t)$, with $q(x_t|x_{t-1})$ the forward noising process. The framework's contribution is the decision-conditioning loop: a frozen production scorer $\pi$ scores each recovered candidate $\hat{x}_0$, and a binary cross-entropy decision penalty $\mathrm{BCE}(\pi(\hat{x}), y)$ feeds back into the trajectory, steering generation by the decision the scorer makes on each candidate rather than by distributional fit alone. Decoding yields an unlimited, privacy-safe synthetic customer base. (b) Strategy Simulation Fidelity (SSF), the share of campaign-budget decisions that can be safely delegated to the synthetic system. Across a family of candidate campaigns, each column pairs the go/no-go decision taken on the real population (top) with the one taken on the synthetic population (bottom); green marks agreement (safe to delegate), red a divergence that puts budget at risk and is routed to human review. SSF is the agreement share, read against a four-zone band from untrusted to fully trusted, with the deployment-ready threshold the production target.

These three gaps cut across communities not previously brought into dialogue, and producing all three artefacts within one framework, SSF for the first, the PolicySynth framework and its evaluation for the second, and the three-axis standard for the third, each tested across two structurally different decision problems, is the principal methodological contribution of this work.

## 3. Strategy simulation fidelity

A synthetic-data-driven decision support system earns adoption only if the decision-maker can certify one thing: that the recommendations it produces on the synthetic population are the recommendations they would have reached on the real one. No established synthetic-data criterion measures that property directly. This section makes it the object of evaluation: we name the quantity to be minimised the *decision-alignment gap*, give its operational measure, *strategy simulation fidelity* (SSF), and attach to SSF a four-zone band that turns it into a rule for when a synthetic-data layer may drive budget decisions. Panel (b) of Figure 1 shows the construct: real-data and synthetic-data go/no-go decisions placed side by side across a family of candidate campaigns, with SSF the share on which they agree.

### 3.1. The decision-alignment gap

A synthetic-data-driven DSS generates a population, applies a strategy to it, and recommends an action; it is useful exactly when that action matches the one a decision-maker would have taken on the real population. The decision-alignment gap is the probability, over a parameterised family of candidate strategies, that the two actions differ. Treating decision agreement rather than statistical resemblance as the property to certify carries to the evaluation of synthetic-data layers a principle established on the training side of the predict-then-optimise literature, that a model is judged by the decisions it induces, not by a generic predictive loss [13, 41, 42, 43]; the decision-alignment gap is the criterion that principle implies for the data layer.

### 3.2. Setting and notation

Let $D_{\text{real}}$ be a real population of $n$ records from a target distribution $P^\star$ over a mixed-type feature space $\mathcal{X}$, and $D_{\text{synth}}$ a synthetic population of $m$ records from the learned distribution $\hat{P}$ of a generator fitted to $D_{\text{real}}$. A decision-support scorer $\pi : \mathcal{X} \to [0, 1]$ estimates each customer's probability of the decision-relevant outcome (churn in retention, subscription in acquisition), and a value function $v : \mathcal{X} \to \mathbb{R}_{\geq 0}$ estimates each customer's monetary worth. Together they turn raw records into the inputs a campaign decision requires.

### 3.3. Strategy simulation fidelity

A strategy $S$ maps a population to a binary go/no-go decision, $S(D) = \mathbf{1}[V_S(D) > 0]$, where $V_S(D)$ is the expected value of executing $S$ on $D$. For a campaign that targets the customers a scorer threshold flags within a segment, $V_S$ sums the expected per-customer return net of contact cost over the flagged set $\mathcal{T}(D, S)$,

$$V_S(D) = \sum_{i \in \mathcal{T}(D,S)} \left[ \pi(x_i)\, r\, v(x_i)(1 - d) - \kappa \right],$$

with success rate $r$, incentive $d$, and per-customer contact cost $\kappa$; the retention and acquisition instantiations and all constants are given in Supplementary Materials (Appendix A and Table S1). A strategy is thus the expected-value decision rule that simulation-based decision support has used since its origins to convert a modelled population into a recommendation [16, 17, 44, 45].

What is new here is that the population is itself learned, so the rule inherits the synthetic layer's decision-alignment error. Each strategy is parameterised by a target segment, an incentive level, a scorer-probability eligibility threshold, and a horizon multiplier; their cartesian product is the strategy family $\mathcal{F}$, instantiated to $K = 560$ strategies on the telecommunications corpus and $K = 600$ on the banking corpus.

**Definition** (Strategy Simulation Fidelity, SSF)**.** *Given a strategy family $\mathcal{F} = \{S_1, \ldots, S_K\}$, the strategy simulation fidelity of a synthetic population $D_{\text{synth}}$ relative to a real population $D_{\text{real}}$ is*

$$\text{SSF}(D_{\text{real}}, D_{\text{synth}}; \mathcal{F}) = \frac{1}{K} \sum_{S \in \mathcal{F}} \mathbf{1}\left[ S(D_{\text{real}}) = S(D_{\text{synth}}) \right].$$

SSF is the share of campaigns on which the synthetic and real populations lead to the same decision; its complement, the decision-alignment gap, is $1 - \text{SSF}$. An SSF of 1.0 is perfect alignment and 0.5 is no information beyond chance.[1] The scale beneath the decision rows in Figure 1(b) reads SSF as a four-zone deployment authority:

- *Untrusted* (SSF $\approx$ 0.5): no information beyond chance; recommendations are not safe to act on.
- *Partial trust* (SSF indicatively $[0.65, 0.85]$): signal is present but acceptable only for exploratory ranking, not budget commitment.
- *Deployment-ready* (SSF $\geq$ 0.90, the production target throughout): may drive go/no-go recommendations the team commits budget against, with the residual disagreements routed to human review.
- *Fully trusted* (SSF $\to$ 1.0): real-data decisions reproduced exactly; no human-review residual.

As a procurement criterion, SSF approves a candidate synthetic-data tool for production only above the deployment-ready threshold and prescribes a human-review queue for the residual. SSF is, finally, consistent with distributional accuracy: for any family with bounded $V_S$, the gap $1 - \text{SSF}$ vanishes as the total-variation distance between $P^\star$ and $\hat{P}$ tends to zero, so distributional closeness is necessary for decision faithfulness. The quantitative bound is left to future work, and Section 5 reports the empirical TVD–SSF relationship [46, 47, 48, 49].

[1] The estimator $\widehat{\text{SSF}}$ is unbiased with variance at most $1/4K$. We report bootstrap 95% confidence intervals (2,000 resamples) for single-seed estimates and means with standard deviations across five training seeds for the multi-seed results that constitute the primary findings.

### 3.4. Why SSF is not redundant with TSTR

A natural objection is that SSF duplicates train-on-synthetic-test-on-real $\Delta$AUC (TSTR), the standard downstream quality metric [30]. It does not. TSTR scores the ranking of customers by predicted risk; SSF scores the go/no-go decisions taken once that ranking is converted into a return calculation. The two can diverge:

**Proposition.** *A generator may achieve low TSTR $\Delta$AUC, preserving the predictive ranking of churn risk, while exhibiting a substantial decision-alignment gap, if its calibration of $\pi$ is systematically biased near the decision threshold $V_S \approx 0$; conversely, a worse-ranking but better-calibrated generator may achieve higher SSF.*

The mechanism is calibration at the threshold. TSTR averages over the whole score range, whereas $S(D) = \mathbf{1}[V_S(D) > 0]$ depends only on the sign of $V_S$, which a threshold-local calibration error can flip without changing the ranking. Because modern probabilistic classifiers are routinely miscalibrated even when their ranking is accurate [50, 51], this regime is the rule, not the exception, and it is exactly where the two metrics part company. We therefore rest the proposition on this mechanism and report SSF and TSTR jointly rather than substituting one for the other. Section 5 quantifies the divergence empirically across the competitive generators.

### 3.5. Contested-zone decomposition

Not every decision discriminates between generators. A campaign whose real-data return is overwhelmingly positive or negative is decided correctly by any competent population; the synthetic layer's accuracy matters only where reasonable decision-makers would hesitate. Because the full-family SSF averages over both kinds, we partition $\mathcal{F}$ by the magnitude of $V_S(D_{\text{real}})$ against a contested-band threshold $\tau$: clearly profitable ($V_S > \tau$), contested ($|V_S| \leq \tau$), and clearly unprofitable ($V_S < -\tau$). We set $\tau$ per corpus to the 33rd percentile of $|V_S(D_{\text{real}})|$ (USD 5,000 on telecommunications, EUR 924 on banking); Section 5.5 confirms the between-generator ranking is stable as $\tau$ sweeps the 25th to 50th percentile of $|V_S|$ (Table S5). The clearly-resolved strategies are recovered by any reasonable population; the contested zone is where a good and a poor synthetic layer separate, and contested-zone SSF is reported separately throughout Section 5 as the discriminating diagnostic.

### 3.6. The three-axis reporting standard

Decision alignment certifies that the layer produces the right recommendations, but deployment requires two properties it does not measure: the synthetic population must not leak the identities or attributes of the real customers behind it, and its records must not be verbatim copies, which offer no protection at all. We therefore report decision alignment jointly with privacy resistance, measured by membership-inference-attack AUC, a value near 0.5 indicating an attacker cannot distinguish training records [52, 53, 54, 55], and record novelty, the share of synthetic records that copy no real record, a value near 100% being required under data-protection regulation [56, 57, 58, 59]. The deployment triple is (SSF, MIA-AUC, novelty): no axis suffices alone, and Section 5 shows a bootstrap baseline that matches PolicySynth on SSF yet fails both other axes categorically, which is the empirical case for reporting all three.

## 4. The PolicySynth decision support framework

Existing tabular generators optimise distributional objectives that, as Section 3 establishes, do not measure decision alignment, and none is designed against the SSF criterion. PolicySynth fills that gap: a pipeline that turns a real customer population into a privacy-protected synthetic one whose go/no-go recommendations, across a parameterised strategy family, agree with those that would be made on real data. Figure 1(a) shows the framework end to end. Such a layer must satisfy three requirements, and they are not of equal research standing.

### 4.1. Design requirements

Drawing on the synthetic-data-for-DSS literature [19, 20, 21] and on production retention systems, we require: **(R1) decision fidelity**, go/no-go recommendations on the synthetic population must agree with those on the real population; **(R2) segment preservation**, the high-value customer subdistribution must be accurate enough for downstream segment analytics; and **(R3) adjustable privacy**, protection against re-identification must be tunable. Only R1 requires an architectural innovation, the decision-conditioning loop, which is the contribution this paper claims and the only mechanism held to the ablation standard of Section 5. R2 and R3 are met by deployment engineering, customer-value stratification and tiered differential privacy, which we adopt and evaluate against the operational criteria they serve. We treat R1 in depth and R2, R3 as deployment considerations, in proportion to the claims made for each.

### 4.2. Architecture: diffusion with decision-conditioning

The architecture is a tabular denoising diffusion model augmented with a decision-conditioning loop. The backbone is standard and serves as setting; the loop is the contribution.

*Backbone.* Each record is encoded into a continuous-relaxed vector (one-hot categoricals, scaled continuous features) following TabDDPM [27]; $d = 41$ on telecommunications and $d = 48$ on banking (Section 5.7). The generative core is a standard $\epsilon$-prediction diffusion model ($T = 100$ steps, linear $\beta$-schedule) trained on the usual denoising objective $\mathcal{L}_{\text{diff}}$ [27]. We choose diffusion over GAN and VAE alternatives for one deployment-driven reason: the seed-to-seed training stability that consistent monthly recommendations demand, and that Section 5 confirms it provides.

*Decision-conditioning.* A model trained on $\mathcal{L}_{\text{diff}}$ alone reproduces the joint distribution but is not steered toward decision fidelity: a population whose churn probabilities drift systematically from the real ones mis-costs every campaign evaluated on it, however well it matches the data otherwise. The requirement is not “match the marginal of churn” but “produce records the production scorer evaluates as it would the real records they replace.” PolicySynth meets it by coupling the backbone to a differentiable surrogate $f_\phi$ of the production scorer $\pi$ (the deployed gradient-boosted churn model, AUC 0.844, which defines $V_S$ and SSF). Because $\pi$ is non-differentiable, $f_\phi$ is a small MLP trained to reproduce its decisions and then frozen (Appendix B, Table S4). The decision penalty backpropagates through $f_\phi$: at each denoising step it recovers an estimate $\hat{x}_0$, scores it, and penalises disagreement with the intended class $y$,

$$\mathcal{L}_{\text{cond}} = \text{BCE}\big(f_\phi(\hat{x}_0),\ y\big), \mathcal{L} = \mathcal{L}_{\text{diff}} + \lambda\, \mathcal{L}_{\text{cond}}, \lambda = 0.5.$$

The coral path in Figure 1(a) is this gradient: the denoising network never sees the scorer, only the gradient of the penalty it induces, which steers each step toward records the scorer reads as it reads real records of the same class. Conditioning on the deployed scorer rather than a fresh one fitted at synthesis is deliberate: it aligns the synthetic layer with the exact model whose decisions it must support and keeps synthesis auditable against a known production artefact, at the cost of a coupling to scorer-retraining cycles. Section 5 isolates this mechanism by ablation and bounds the regime in which it is load-bearing.

### 4.3. Deployment considerations

Two further mechanisms address R2 and R3. Neither is claimed as a methodological contribution; both are deployment engineering, reported because a deploying organisation must understand them.

*Segment preservation (R2).* Uniform sampling under-represents the high-value tail, so PolicySynth weights top-decile customer-lifetime-value records by 2.0 in minibatch construction. We initially proposed this as a contribution to decision alignment; the factorial ablation of Section 5 does not support that on telecommunications, so we report it honestly as a deployment feature. The reclassification is instructive: not every plausible mechanism improves decision alignment, and SSF is what tells the two cases apart.

*Adjustable privacy (R3).* Because regulation distinguishes feature classes, PolicySynth assigns each feature to one of three sensitivity tiers ($\varepsilon_1 = 0.5$, $\varepsilon_2 = 2.0$, $\varepsilon_3 = 8.0$) under tier-wise DP-SGD. We claim no contribution to the differential-privacy literature: the composed budget is weak-to-moderate ($\varepsilon_{\text{total}} \leq 10.5$), and the privacy property the framework actually claims is the empirical membership-inference resistance reported in Section 5. The operational value is the per-tier allocation, which lets a team tighten any tier without retraining; a stronger formal bound via Rényi accounting ($\varepsilon \approx 2$–$4$) is left to future work.

### 4.4. Production variant: marginal anchoring

Some environments must pass a strict marginal distribution audit before approval, even though such an audit is not by itself evidence of decision faithfulness. The generative core reproduces the joint well in expectation but leaves residual drift on a few categorical marginals; for these settings we document a transparent post-processing step, marginal anchoring, that reweights the synthetic population by the real joint frequencies of four anchor variables, yielding the variant PolicySynth-A. The trade-off it exposes is itself a finding (Section 5): anchoring passes a strict ±3-point marginal gate and nudges SSF upward, but worsens the ROI gap by distorting the joint structure that carries the decision signal. Distributional accuracy on auditable marginals and decision faithfulness on ROI magnitudes can move in opposite directions, and a team that optimises one without measuring the other can erode the very decision agreement the layer was deployed to provide.

# 5. Empirical evaluation and deployment case study

The central empirical result of this paper is a stability finding. Across repeated retraining, PolicySynth's go/no-go recommendations move by roughly a tenth of the amount the strongest competing neural generator's do. It is this seed-to-seed stability, not peak single-run accuracy, that determines whether a synthetic-data layer can be trusted to drive recurring budget decisions, and the evaluation that follows is built to expose it.

We establish the result in seven steps, each tied to a research question of Section 1. Section 5.1 fixes the experimental design and the two corpora. On the telecommunications corpus, Section 5.2 reports the stability finding itself (RQ1); Section 5.3 attributes it by factorial ablation to a single architectural mechanism and bounds the conditions under which that mechanism matters (RQ2); Sections 5.4 and 5.5 show that the decision-alignment criterion is neither redundant with the standard downstream metric nor an artefact of the evaluation's scorer and threshold choices (RQ1); and Section 5.6 hardens the criterion into a three-axis deployment gate (RQ3). Section 5.7 then replicates the entire evaluation on a structurally different banking corpus, separating what is a property of the framework from what is a property of one decision problem. Section 5.8 instantiates the framework as an operational DSS and develops the managerial implications.

## 5.1. Experimental design

The primary corpus is the IBM Telco Customer Churn benchmark: 7,043 subscribers described by 20 behavioural and contractual features, with a 26.5% churn rate that places roughly USD 5.3 million of 24-month customer lifetime value at risk. We partition it 75/25 into training ($n$ = 5,282) and holdout ($n$ = 1,761), stratified on churn (Table S2). The corpus was chosen for three properties that make it a discriminating test rather than a convenient one: (i) a realistic mixed-type structure that exercises the encoding; (ii) a non-trivial joint dependency (service add-ons are conditional on internet service) that separates generators which model the joint from those which only match marginals; (iii) and full public availability, so that every number in this section is independently reproducible. To distinguish framework properties from decision-problem properties, we replicate the full pipeline on the UCI Bank Marketing corpus in Section 5.7: a different decision (term-deposit acquisition), a different base rate (11.5%), and a different value function (Table S2).

The strategy function $S$ depends on a churn-probability scorer $\pi$: a gradient-boosted-tree classifier (200 trees, depth 4, learning rate 0.05) with holdout AUC 0.844 (Table S3). This same scorer is the frozen production scorer inside PolicySynth's decision-conditioning loop, which raises an obvious objection that the framework's SSF advantage might be an artefact of conditioning on the very model used to evaluate it. We confront this rather than defer it: Section 5.5 shows the advantage persists under scorers of materially different inductive bias that never entered the loop and, more tellingly, that PolicySynth scores highest under a held-out scorer it was never conditioned on.

The family $\mathcal{F}$ comprises 560 strategies parameterised by four target segments × ten discount levels × seven eligibility thresholds × two campaign horizons. This is the strategy family against which every SSF, contested-zone, factorial-ablation, and scorer-sensitivity result reported in this section is measured.

We compare four generative families (CTGAN, TVAE, TabDDPM, REaLTabFormer) against PolicySynth, plus a bootstrap resample-with-replacement control. The bootstrap is included as a deliberately double-edged baseline: a perfect-information control on the distributional axis, since it reproduces marginals exactly, and a no-information control on privacy, since every record it emits is a verbatim real customer. It is the sharpest available test of whether decision alignment alone can certify a synthetic-data layer and, as Section 5.6 shows, it cannot.

## 5.2. The stability finding: decision alignment across retraining

Decision alignment that cannot be reproduced across retraining cycles is not deployable, however high its peak value. The headline result of this paper is that PolicySynth's decision alignment is an order of magnitude more stable across training seeds than the strongest neural competitor's, and that this stability is the property on which a synthetic-data layer for managerial use must be judged (RQ1). Table 1 reports the multi-seed evidence; Figures 2 and 3 display it.

PolicySynth attains a mean SSF of 0.923 across five seeds at a standard deviation of 0.012. CTGAN attains a near-identical mean (0.884) at a standard deviation of 0.115. Figure 3 makes the contrast unmistakable: PolicySynth's five seeds fall in a band of width 0.03 ([0.91, 0.94]), whereas CTGAN's span [0.69, 0.98], from below the partial-trust floor to near-perfect agreement. Two generators thus report almost the same average decision alignment while offering entirely different guarantees about next month's retraining cycle.

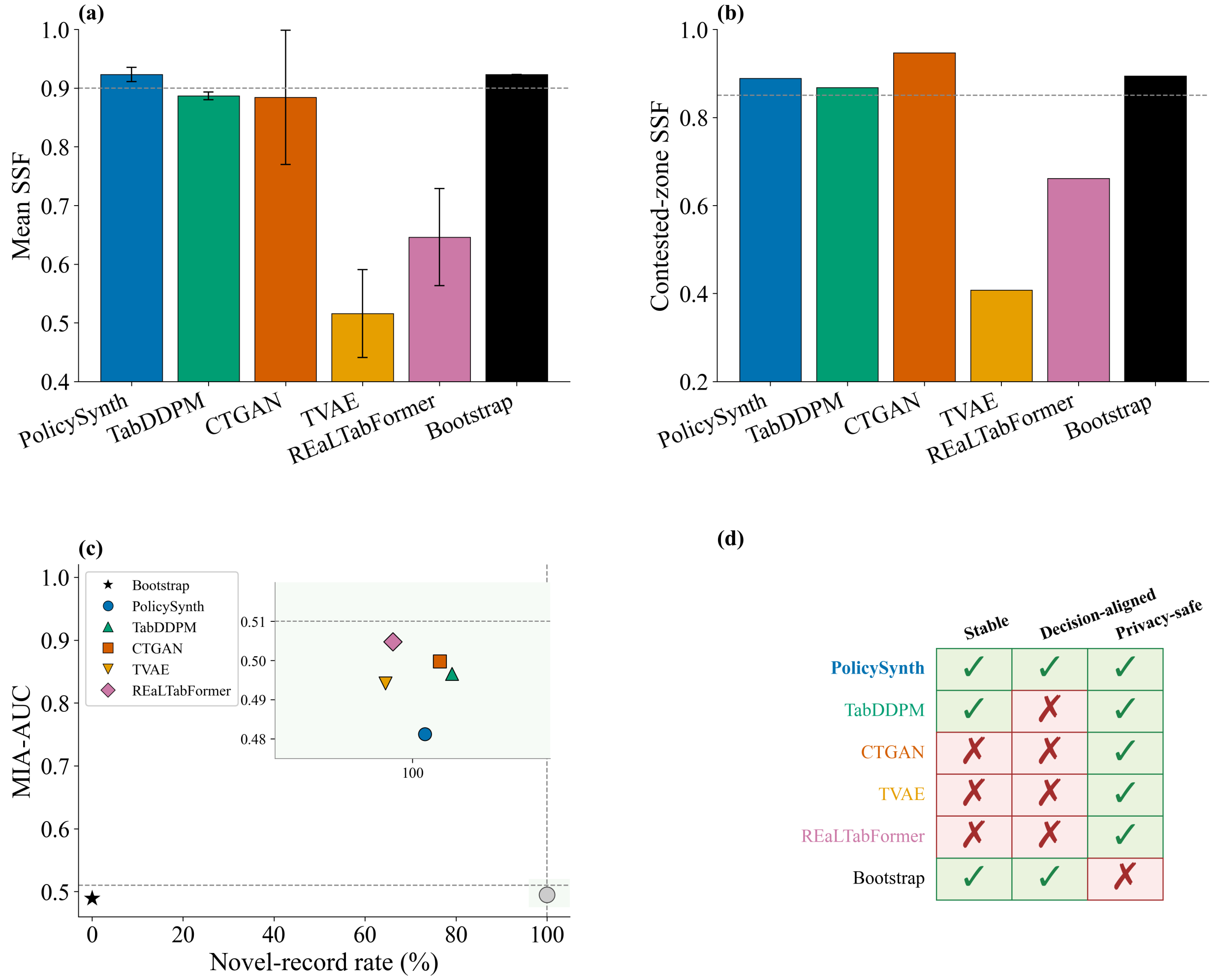

**Figure 2:** Decision alignment, stability, and privacy on the telecommunications corpus (560-strategy family). (a) Mean SSF across five seeds; error bars span ±1 SD. PolicySynth and TabDDPM clear or approach the 0.90 deployment line with tight bars; CTGAN's near-equal mean hides an order-of-magnitude wider spread. (b) Contested-zone SSF (canonical seed, $n = 189$); CTGAN's single-seed peak (0.947) does not survive the seed variance shown in panel (a). (c) Privacy plane: novel-record rate versus MIA-AUC. The lower-right region is privacy-safe; PolicySynth holds the lowest MIA-AUC (0.481), while the bootstrap sits alone at 0% novelty. (d) Deployment gate. Stable: SSF SD $\leq 0.02$; Decision-aligned: mean SSF $\geq 0.90$ and contested-zone SSF $\geq 0.85$; Privacy-safe: MIA-AUC $\leq 0.51$ and novelty = 100%. PolicySynth alone clears all three.

Because SSF is bounded and non-normal at five seeds, we test the variance difference with a non-parametric Fligner-Killeen test and a bootstrap confidence interval on the standard-deviation ratio, rather than the variance-ratio F-test that assumes normality. The bootstrap 95% CI for the SD ratio (CTGAN over PolicySynth) is [1.90, 27.47], bounded entirely above parity. The Fligner–Killeen statistic is $p = 0.16$: the expected low power of a five-sample variance test, not a weak effect. On the banking corpus, where fifteen seeds give adequate power, the same comparison returns $p = 0.004$ (Section 5.7). The stability advantage is therefore established by a ratio interval that excludes parity on both corpora and reaches conventional significance once the seed count is sufficient.

The managerial translation is direct. A standard deviation of 0.012 means an organisation that retrains its PolicySynth-based DSS monthly can expect its go/no-go recommendations to move by at most ±1.2 percentage points between cycles; the same organisation on CTGAN faces swings of ±11.5 points: on the order of a reversed recommendation for one campaign in nine between consecutive runs. A simulator that revises its advice that often on unchanged inputs cannot anchor a budget-approval workflow, however well it scores on its best seed.

A final reading of Table 1 anticipates the rest of the section. PolicySynth is the only method that clears the 0.90 deployment threshold on the mean while also satisfying the SD $\leq$ 0.02 stability criterion. TabDDPM is stable but sits below the alignment threshold (0.887); CTGAN and the weaker generators fail stability; and the bootstrap control matches PolicySynth on SSF exactly (0.923) while failing the privacy axis outright. Panel (d) of Figure 2 summarises these pass/fail outcomes against the full deployment gate. Per-seed values are reported in Supplementary Materials (Table S3).

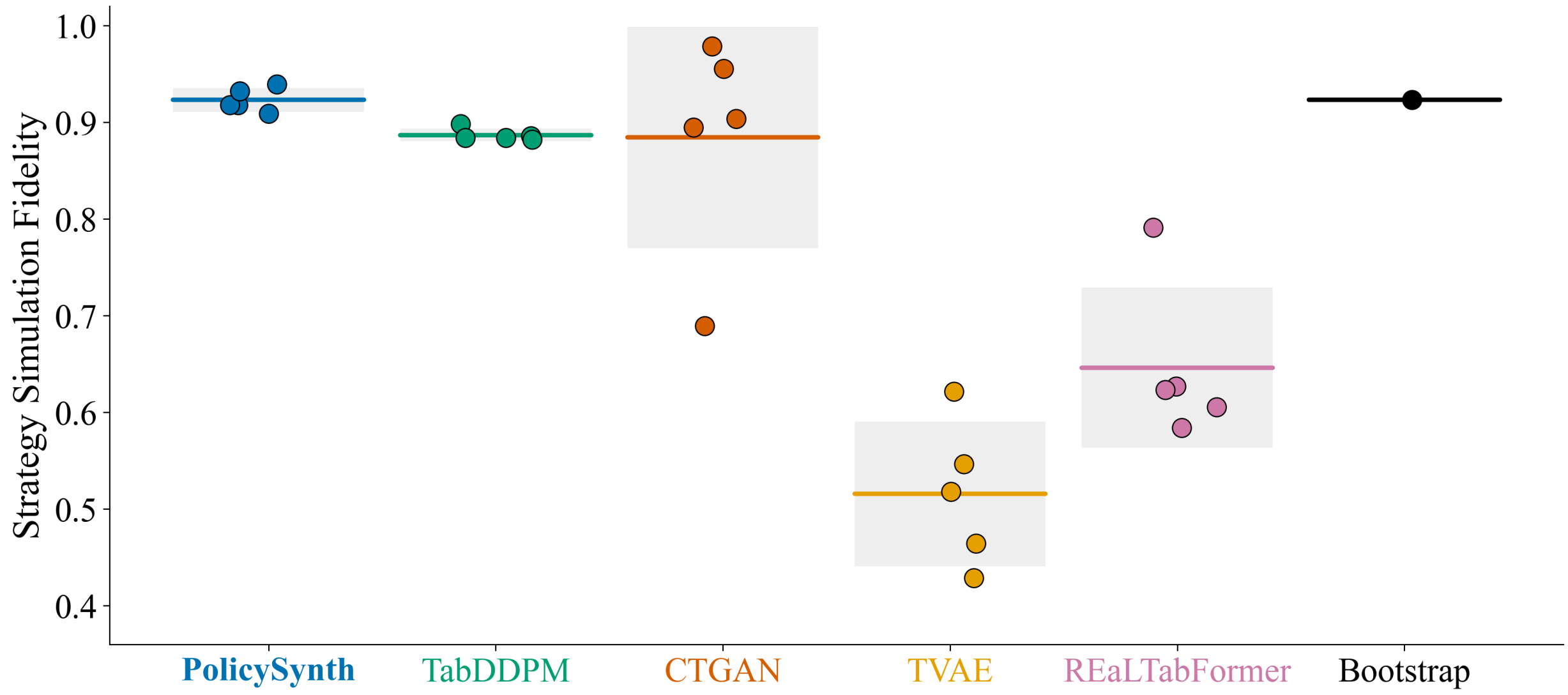

**Figure 3:** Seed-to-seed decision-fidelity stability on the telecommunications corpus. Each marker is the SSF of one training seed; the coloured line is the seed mean and the grey band spans $\pm 1$ SD. PolicySynth's five seeds occupy a band of width 0.03 ($[0.91, 0.94]$); CTGAN's span $[0.69, 0.98]$, from below the partial-trust floor to near-perfect. Bootstrap is a single point (deterministic given the outer seed; SD $= 0$ by construction); its high SSF is matched by the privacy failure of Figure 2(c). For an organisation retraining monthly, PolicySynth's narrow band means recommendations hold across refresh cycles; CTGAN's width means the same campaign can flip between cycles.

**Table 1**
Multi-seed decision alignment and joint quality on the $K = 560$ strategy family (mean $\pm$ sample SD across five seeds; MIA-AUC and novel-record rate from the canonical seed). PolicySynth is the only method that is simultaneously decision-aligned (mean SSF $\geq 0.90$), stable (SD $\leq 0.02$), and privacy-safe.

| Method | Mean SSF (agreement) | SSF SD (stability) | Mean ROI gap (error, %) | Mean KS (distr.) | MIA-AUC (privacy) | Novel-record rate |
|---|---|---|---|---|---|---|
| **PolicySynth** | **0.923** | **0.012** | **107** | **0.148** | **0.481** | **100 %** |
| TabDDPM | 0.887 | 0.007 | 111 | 0.161 | 0.497 | 100 % |
| CTGAN | 0.884 | 0.115 | 163 | 0.295 | 0.500 | 100 % |
| REaLTabFormer | 0.646 | 0.083 | 318 | 0.838 | 0.505 | 100 % |
| TVAE | 0.516 | 0.075 | 124 | 0.367 | 0.494 | 100 % |
| Bootstrap$^{\dagger}$ | 0.923 | 0.000 | 113 | 0.019 | 0.489 | **0 %** |

## 5.3. Architectural attribution: factorial ablation

The framework's stability traces to a single architectural mechanism, not to the joint action of several. A 2×2 factorial ablation of the two mechanisms that distinguish PolicySynth from a plain tabular diffusion model (decision-conditioning ($A_2$) and customer-value stratified sampling ($A_3$)) isolates the cause of its decision-alignment performance and, with the banking replication, bounds the conditions under which each mechanism matters (RQ2). Table 2 reports the four configurations on the 560-strategy family.

Decision-conditioning is the load-bearing mechanism. Removing it alone reduces canonical-seed SSF by 12.2% points ($0.927 \rightarrow 0.805$) and widens the ROI gap by 35 points. The effect is an order of magnitude larger than the full framework's multi-seed standard deviation ($\pm 0.012$), so it cannot be a finite-seed artefact: decision fidelity is not preserved by the diffusion backbone alone, and the framework's headline performance traces to this one mechanism rather than to a combination. Customer-value stratification, by contrast, does not improve decision alignment on this corpus, disabling it leaves SSF marginally higher (0.938 versus 0.927), so we reclassify it honestly as a deployment feature for segment-distribution preservation rather than a contribution to decision fidelity. That SSF cleanly separates a mechanism that matters from one that does not is itself evidence for the criterion.

**Table 2**
Factorial ablation of decision-conditioning ($A_2$) and customer-value stratification ($A_3$). All four configurations are trained from the canonical seed; SSF and ROI gap are computed on the holdout against the same scorer.

| Configuration | $A_2$ (dec.-cond.) | $A_3$ (CLV strat.) | SSF | ΔSSF vs full | ROI gap (%) |
|---|---|---|---|---|---|
| Full PolicySynth | ✓ | ✓ | 0.927 | — | 96 |
| Decision-conditioning ablated | × | ✓ | 0.805 | −0.122 | 131 |
| CLV-stratification ablated | ✓ | × | 0.938 | +0.011 | 95 |
| Both ablated | × | × | 0.855 | −0.072 | 110 |

The banking replication shows that this attribution is conditional and predictably so (Figure 4b). Where the decision is governed by a single high-variance value magnitude rather than by a label-correlated joint, removing decision-conditioning moves SSF by only +0.2 points because the diffusion backbone already reproduces the deciding magnitude. We report the ablation throughout as $\Delta\text{SSF} = \text{SSF}_{\text{without}} - \text{SSF}_{\text{full}}$, so a load-bearing mechanism is negative and a redundant one is near zero. Decision-conditioning is therefore load-bearing precisely when the go/no-go decision turns on the joint between the outcome label and the covariates the scorer reads, and redundant when a value magnitude the backbone reproduces dominates. This is not a description fitted after the fact to two datasets but a prediction the architecture makes: a deployment team can determine which regime it faces before training, from whether its value function is dominated by one high-variance magnitude or by a label–covariate interaction. RQ2 is thus answered conditionally, with the condition stated in advance.

## 5.4. Decision alignment is not redundant with TSTR

Among the generators a practitioner would actually deploy, the standard downstream-quality metric carries no information about decision alignment. Train-on-synthetic-test-on-real ΔAUC (TSTR) and SSF are statistically unrelated across the competitive set, the empirical content of Proposition 1 and the reason the decision-alignment criterion is not redundant with the existing toolbox (RQ1).

We evaluated eighteen (method, seed) pair: six methods across three seeds with fully reproducible synthetic-population caches, on both SSF and TSTR ΔAUC. Across all six methods the Pearson correlation between (1 − SSF) and ΔAUC is strong ($r = +0.874$, $p = 2.2 \times 10^{-6}$, $n = 18$). But that correlation is produced entirely by the two broken generators that fail on both axes at once (REaLTabFormer, mean SSF ≈ 0.65; TVAE, ≈ 0.52): TSTR correctly ranks generators that are bad everywhere. Restricted to the four competitive methods a deployment team would actually consider (bootstrap, CTGAN, TabDDPM, PolicySynth) the correlation collapses to $r = -0.140$ ($p = 0.66$, $n = 12$): flat across the competitive band, the small negative coefficient indistinguishable from noise. The full scatter over the 18 (method, seed) pairs is Figure S1; the correlation procedure is in Supplementary Materials (Appendix C).

We therefore rest Proposition 1 on its mechanism: miscalibration near the decision threshold flips the sign of $V_S$ without materially changing the churn ranking TSTR rewards, and report the restricted-set result as consistent, if underpowered, corroboration rather than a decisive test. The operational consequence is unambiguous: a team that certifies a synthetic-data layer on TSTR alone has no signal on whether it will reproduce real-data decisions, precisely in the regime of competent generators where the choice between candidates is actually made.

## 5.5. Robustness across scorers and the contested zone

The decision-alignment result survives the two evaluation choices a sceptic questions first and the scorer analysis additionally turns the circularity concern of Section 5.1 on its head (RQ1).

SSF is computed through a downstream scorer $\pi$, so we recompute it under three scorers of materially different inductive bias: gradient-boosted trees (holdout AUC 0.844), logistic regression (0.847), and a multilayer perceptron (0.786). Table 3 reports the result. The ordering of the four competitive methods is preserved across all three scorers, though absolute values shift by up to 13 points between the gradient-boosted and perceptron scorers, which is why we recommend, as general evaluation practice, that SSF always be reported under at least two scorers of different inductive bias, each named with its holdout AUC. Two readings of Table 3 matter for the present claims. First, the table reports canonical-seed values, on which CTGAN's SSF marginally equals or exceeds PolicySynth's under every scorer; as Section 5.2 establishes, this single-seed ordering does not survive seed variance (CTGAN SD = 0.115 against PolicySynth's 0.012), and the deployment-relevant comparison is the multi-seed one. Second, the advantage is not an artefact of the conditioning signal: PolicySynth retains high SSF under independently trained scorers of materially different inductive bias that never entered the loop, logistic regression (0.830) on telecommunications and held-out random-forest and logistic-regression scorers on banking (SSF 0.95 and 0.98), none of which shares the MLP form of the surrogate $f_\phi$ (Table S5 and S6). The TVAE row illustrates the diagnostic in the other direction — its 51-point swing across scorers (0.429 to 0.939) exposes a distributional fragility consistent with its multi-seed instability (Table 1, SD = 0.075).

**Table 3**
Scorer sensitivity of canonical-seed SSF on the $K = 560$ family. (GBM: gradient-boosted trees, LogReg: logistic regression, MLP: multilayer perceptron.)

| Method | GBM SSF (AUC 0.844) | LogReg SSF (AUC 0.847) | MLP SSF (AUC 0.786) | Range |
|---|---|---|---|---|
| PolicySynth | 0.918 | 0.830 | 0.964 | 0.134 |
| CTGAN | 0.979 | 0.850 | 0.984 | 0.134 |
| TabDDPM | 0.886 | 0.818 | 0.954 | 0.136 |
| TVAE | 0.429 | 0.704 | 0.939 | 0.510 |

**Table 4**
Contested-zone decomposition of canonical-seed SSF (contested zone $|V_S| \leq$ USD 5,000, the 33rd-percentile band defined in Subsection 5.5).

| Method | Contested-zone SSF ($n = 189$) | Clearly profitable SSF ($n = 322$) | Clearly unprofitable SSF ($n = 49$) |
|---|---|---|---|
| **PolicySynth** | **0.889** | 0.922 | 1.000 |
| CTGAN | 0.947 | 0.994 | 1.000 |
| TabDDPM | 0.868 | 0.879 | 1.000 |
| Bootstrap | 0.894 | 0.929 | 1.000 |
| REaLTabFormer | 0.661 | 0.512 | 1.000 |
| TVAE | 0.407 | 0.354 | 1.000 |

Not every campaign decision is equally informative about a generator. A campaign whose real-data return is overwhelmingly positive or negative is approved or rejected correctly by any competent synthetic population; generators separate only where reasonable decision-makers would hesitate. We therefore partition the family by $|V_S(D_{\text{real}})|$ at a contested-band threshold $\tau$ set per corpus to the 33rd percentile (USD 5,000 on telecommunications) yielding 189 contested and 371 clearly-resolved strategies; a sweep of $\tau$ from the 25th to the 50th percentile leaves the between-generator ranking unchanged, so the decomposition is not a threshold artefact. Table 4 reports the partitioned canonical-seed results. In the contested zone the separation between methods is at its widest: PolicySynth's contested-zone SSF of 0.889 is 0.228 above REaLTabFormer's 0.661, whereas in the clearly-resolved zone every method exceeds 0.879 on the clearly-profitable subset and reaches 1.000 on all 49 clearly-unprofitable strategies (Table S7). Errors concentrate exactly where the partition predicts: of the 46 strategies on which canonical-seed PolicySynth disagrees with the real-data benchmark, 21 (46%) lie in the contested zone, which is only 34% of the family. The contested zone matters not because any single near-threshold mis-decision is individually costly but because it is the only region in which competitive generators separate at all, and because near-threshold campaigns are numerous enough that small, repeated mis-commitments accumulate into a budget-discipline and trust cost the full-family average conceals. It is, in short, the region where a decision-faithful synthetic-data layer earns its cost, which is why we report it separately throughout.

### 5.6. The three-axis reporting standard

A single evaluation axis can always be gamed, and the bootstrap control proves it. Resampling real records with replacement attains a canonical-seed SSF of 0.923 — statistically indistinguishable from PolicySynth's 0.918 (paired permutation test over the 560-strategy family, $\Delta = -0.005$, $p = 0.63$, 5,000 permutations) — and is, on decision alignment alone, the cheapest recommendable choice: no training, deterministic output, trivial to implement. Were SSF the sole reported criterion, a deployment team would rationally select it. Yet every record a bootstrap emits is a verbatim real customer: its novel-record rate is 0%, and any membership-inference attack that compares synthetic records against the real corpus by exact match returns AUC = 1.0 (the distance-based attack reported in Table 1, 0.489, understates the failure because it never registers an exact duplicate). The bootstrap is therefore categorically disqualified for any DSS under privacy regulation, whatever its SSF. Panel (c) of Figure 2 isolates it at the privacy-failing extreme of the novelty axis, alone outside the cluster of generative methods at 100% novelty, a cluster in which PolicySynth additionally posts the lowest membership-inference AUC (0.481), the strongest privacy resistance in the comparison.

This counter-example generalises into the section's answer to RQ3. The minimum reporting standard for a synthetic-data layer entering production is the joint triple (SSF, MIA-AUC, novel-record rate), because each single-axis report admits a counter-example: SSF alone passes the privacy-failing bootstrap; MIA-AUC alone passes decision-broken TVAE (MIA = 0.494, within sampling noise of the privacy-perfect 0.500, yet mean SSF = 0.516, a coin toss on the decision); and novelty alone is satisfied by every non-bootstrap method at 100%. Only the three axes read together separate a deployable layer from an undeployable one, the logic Panel (d) of Figure 2 renders as a pass/fail gate, on which PolicySynth alone clears all three columns.

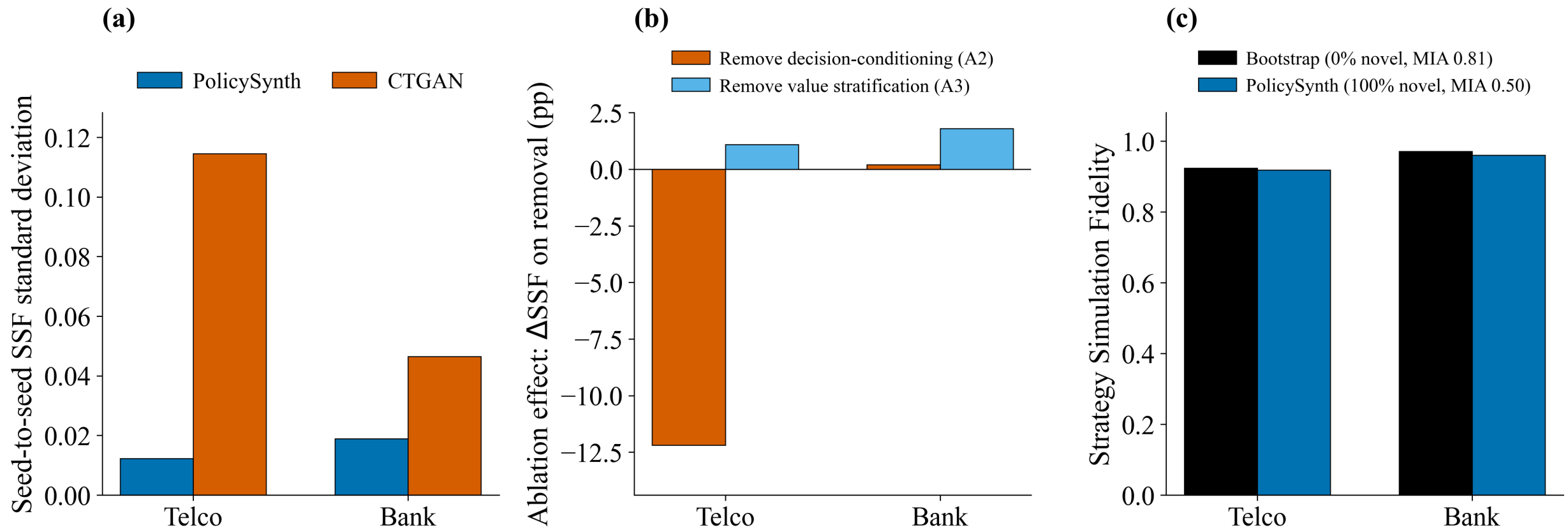

**Figure 4:** Two-corpus generality. (a) Seed-to-seed SSF standard deviation, PolicySynth versus CTGAN, on telecommunications and banking; PolicySynth is tighter on both (0.012 vs 0.115; 0.019 vs 0.047), with the bootstrap SD-ratio 95% CI bounded above parity in each case ($[1.90, 27.47]$; $[1.44, 5.76]$). (b) Ablation effect as $\Delta\text{SSF} = \text{SSF}_{\text{without}} - \text{SSF}_{\text{full}}$, with the other mechanism enabled, so a load-bearing mechanism is negative. Removing decision-conditioning costs −12.2 pp on telecommunications but only +0.2 pp on banking; removing value stratification is near zero on both (+1.1, +1.8). (c) Three-axis counter-example on both corpora: the bootstrap matches PolicySynth on SSF yet fails on novelty (0% vs 100%) and privacy (MIA-AUC 0.81 vs 0.50), confirming the failure SSF alone cannot detect is dataset independent.

## 5.7. Generality: replication on a second corpus

A finding shown on one corpus may be a property of that decision problem rather than of the framework. We therefore replicate the entire evaluation on UCI Bank Marketing: term-deposit acquisition rather than churn retention, an 11.5% base rate, a leakage-free subscription scorer (AUC 0.756, with the post-outcome duration feature excluded from both scorer and value function), and a value function rebuilt from term-deposit margin on customer balance. The strategy family is rebuilt on score quantiles to accommodate the class imbalance ($K = 600$), and the same six methods run through the identical pipeline. Seed counts are reported per analysis: fifteen seeds for the stability and three-axis comparisons, three per cell for the factorial ablation, single-seed for the contested-zone and held-out-scorer diagnostics. Figure 4 reports all three structural findings across the two corpora; two replicate and the third is the characterised scope condition. Full banking results are in Table S8.

*Stability replicates* (Figure 4a). PolicySynth's seed-to-seed SSF standard deviation is again the tighter of the two: 0.019 against CTGAN's 0.047, a 2.5-fold separation, with the bootstrap SD-ratio 95% CI at $[1.44, 5.76]$ and Fligner–Killeen $p = 0.004$ at fifteen seeds. PolicySynth attains a mean SSF of 0.960 on this corpus. The separation is narrower than the order-of-magnitude gap on telecommunications because the whole diffusion family is more stable here (TabDDPM is tighter still at 0.009), since it shows the advantage arises from diffusion training dynamics rather than corpus-specific tuning, and the ratio CI excludes parity on both corpora.

*Attribution is corpus-dependent* (Figure 4b). Removing decision-conditioning costs 12.2 points on telecommunications but only +0.2 on banking, where the balance-weighted value function governs the decision and the backbone already reproduces the balance magnitude; removing value stratification is near zero on both (+1.1 and +1.8). On banking, neither auxiliary mechanism is load-bearing. This is the scope condition of Section 5.3 confirmed on independent data (decision-conditioning matters where the decision turns on a label-correlated joint and is redundant where a value magnitude dominates) and a robustness check rules out the segmentation as its cause: the null effect persists under a job-category strategy family on which full-family SSF is only 0.69.

*The three-axis standard replicates* (Figure 4c). The bootstrap again matches PolicySynth on SSF (0.971) while failing on novelty (0% against 100%) and on the privacy axis (membership-inference AUC 0.81 against 0.50 for every generative method). The privacy failure that decision alignment alone cannot see is dataset-independent, exactly as the three-axis argument predicts.

A held-out-scorer check closes the circularity concern on this corpus too: PolicySynth attains SSF 0.98 under a held-out random-forest scorer (AUC 0.743) and 0.95 under a held-out logistic-regression scorer (AUC 0.730), neither of which entered the decision-conditioning loop. Across the two corpora, then, the stability finding and the three-axis privacy standard generalise, while the architectural attribution follows the scope condition the framework states in advance rather than holding universally.

**Table 5**
Deployment-time decision agreement. Full-family figures are multi-seed means (517 of 560, mean SSF 0.923); contested-zone and deep-dive figures are canonical-seed (168 of 189, SSF 0.889; 3 of 3 archetypes GO). Family-level SSF, privacy, and novelty are in Table 1 and not repeated here.

| Decision metric | Result |
|---|---|
| Directional agreement, full family (multi-seed mean) | 517 of 560 (SSF 0.923) |
| Directional agreement, contested-zone subset | 168 of 189 (SSF 0.889) |
| Deep-dive directional agreement (archetypes A, B, C) | 3 of 3 (GO on all) |
| Mean absolute ROI gap, three deep-dive archetypes | 75.6 % |

**Table 6**
Empirical volume corrections by campaign archetype (size-matched comparison on 1,761 real-holdout customers vs 1,761 PolicySynth-synthesised customers, canonical seed).

| Archetype | Segment | Threshold $p$ | Real flagged | Synth flagged | Correction factor | Bias direction |
|---|---|---|---|---|---|---|
| Broad / low-threshold | mid | 0.30 | 220 | 127 | ×1.73 | under-flag |
| Precision / high-threshold | high | 0.50 | 182 | 46 | ×3.96 | under-flag |
| All-segments | all | 0.70 | 165 | 44 | ×3.75 | under-flag |

### 5.8. Deployment demonstration and managerial implications

The properties established above translate into a deployable system; this subsection demonstrates the framework in an operational workflow and develops the managerial implications. It is an illustration of the evaluated framework, not an independent validation: the case study runs on the same strategy family as the multi-seed evaluation, so its decision-agreement figures are the deployment-time reading of results already reported, not new evidence.

We instantiate PolicySynth as a retention-marketing DSS for a telecommunications operator with 7,043 active customers, a 26.5% churn rate, and roughly USD 5.3 million of 24-month lifetime value at risk. The synthetic-population store refreshes monthly; campaign designers query it with parameters (segment, discount, eligibility threshold, horizon) and receive a go/no-go recommendation, an estimated ROI, a seed-to-seed confidence band, and the active cycle's three-axis quality reading, without any real customer record entering the analytics environment. At an industry-midpoint 5% waste rate on mis-targeted campaigns, such an operator loses about USD 265,000 per quarter to campaigns a decision-faithful simulator could have flagged in advance.

*What the framework delivers, and where it stops.* PolicySynth is reliable on direction and biased on magnitude. Directionally, it agrees with the real-data benchmark on a multi-seed mean of 517 of 560 strategies (mean SSF 0.923) and on 168 of 189 contested-zone strategies (canonical-seed SSF 0.889), on the deployment-ready side of both thresholds (Table 5). On magnitude, it systematically understates ROI: across three named campaign archetypes the absolute ROI gap is 70–78% (mean 75.6%). The mechanism is a systematic under-flagging — synthetic-flagged volumes run 1.7 to 4.0 times below real-flagged volumes at the same threshold (Table 6) — traceable to a synthetic mean churn probability of 0.178 against the real holdout's 0.263.

This magnitude bias is exactly the failure TSTR cannot see. On Strategy 114 (segment = high, discount = 0.25, threshold = 0.30, full horizon), the real data ROI is −USD 184, PolicySynth's estimate is −USD 4,704, and the bootstrap's is −USD 11,606: a 37-fold disagreement between two methods whose TSTR $\Delta$AUC differ by less than 0.005. All three agree on direction (NO-GO), so SSF and the campaign decision are correct; but the magnitudes that would feed a budget allocation are not interchangeable, which is why magnitude must be read through correction rather than trusted raw.

Deployment teams therefore apply the per-archetype volume corrections of Table 6 as post-hoc multipliers ($\approx \times 1.7$ for broad low-threshold campaigns, $\approx \times 3.8$ for precision and all-segment campaigns), re-estimated on a sliding window of completed campaigns to avoid drift. Folding the correction into the generator is the wrong place for it: a calibration experiment that quantile-maps synthetic churn scores onto the real distribution raises SSF by 4.1 points but widens the ROI gap by 330 points, distorting the joint structure that carries the decision signal (Tables S9 and S10).

Three implications follow for deployment teams. First, privacy-safe pre-testing is a genuine workflow gain: the conventional approval flow exposes real customer records to analyst desktops during cost forecasting, whereas PolicySynth-mediated pre-testing returns aggregate decision outputs without touching the real table, with 100% novel records and MIA-AUC 0.481 as the data-protection officer's evidentiary basis (the tiered $\varepsilon \in [8.0, 10.5]$ is supplementary, not a stand-alone legal guarantee). Second, the three-axis gate should govern promotion: a refresh is promoted only if it jointly clears mean SSF $\geq$ 0.90, SSF SD $\leq$ 0.02, MIA-AUC $\leq$ 0.51, and novelty = 100%, with any breach a deployment-pause condition. Third, the synthetic layer is a decision-quality instrument, not a data replacement: it informs decisions taken before the real table is touched but does not replace that table, and its directional-not-magnitude boundary means ROI must be read through the Table 6 corrections, with contested-zone recommendations queued for human review against the real-data baseline.

# 6. Discussion

## 6.1. Principal findings and what is new

Read against the gap the paper set out to close, the findings converge on one reframing: for managerial use, a synthetic-data layer should be judged by the reproducibility of its decisions, not the accuracy of its best run. Two generators with near-identical mean alignment but order-of-magnitude different seed-to-seed variance make this concrete, and, to our knowledge, this is the first time seed-to-seed decision stability has been proposed as the primary acceptance criterion for a synthetic-data-driven DSS.

Two further claims follow, and it is worth marking precisely what is new in each. First, decision alignment is a construct distinct from predictive-ranking quality: among competent generators the standard downstream metric carries no information about it, the mechanism being threshold miscalibration that flips the sign of the value function without disturbing the churn ranking. We rest this on the mechanism and treat the underpowered restricted-set correlation ($r = -0.140$, $n = 12$) as consistent evidence rather than a refutation, because the honest version of the claim is the defensible one. Second, no single quality axis certifies a layer, as a bootstrap baseline that ties PolicySynth on alignment yet reproduces real customers verbatim demonstrates. The constituent axes (utility, privacy, novelty) are familiar from the synthetic-data literature; what is new is the demonstration that a decision-utility axis and the privacy axis can be jointly satisfied, and their consolidation into an operational promotion gate. We claim the gate and the decision-utility axis as contributions, not the multi-axis principle itself.

What ties the findings together is a scope condition stated in advance: decision-conditioning is load-bearing where the decision turns on a label–covariate joint and redundant where a single value magnitude dominates, a regime a deployment team can identify before training. A method that predicts the conditions under which it is unnecessary is more trustworthy, not less, than one claimed to help everywhere.

## 6.2. Implications for theory and practice

For decision support research, the contribution returns the field to its own foundational commitment. The DSS tradition has held since its inception that a support system is measured by the quality of the decisions it enables, not by the fidelity of its internal representations [60, 61, 62]; SSF carries that commitment into the synthetic-data setting, where it has been absent. It also clarifies the relationship to decision-focused, predict-then-optimize learning [13]: that literature trains models so their predictions yield good decisions, whereas SSF evaluates whether a synthetic population yields the same decisions as the real one. The two operate at different loci and compose naturally: a decision-focused model trained on a decision-faithful synthetic layer is the obvious next construction, which we leave to future work.

For practitioners, three implications are immediately actionable. First, the scope condition is a pre-deployment diagnostic: before adopting a decision-conditioned generator, a team can inspect whether its value function is dominated by one high-variance magnitude (in which case a plain diffusion model suffices) or by a label–covariate interaction, in which case decision-conditioning earns its complexity, avoiding both under- and over-engineering. Second, the three-axis gate is a concrete governance instrument: a refresh is promoted to production only if it jointly clears mean SSF $\geq$ 0.90, stability SD $\leq$ 0.02, MIA-AUC $\leq$ 0.51, and 100% novelty, with any breach a documented deployment-pause, a specification a data-protection officer and a campaign owner can both sign. Third, and most consequential in regulated settings, the framework enables privacy-safe pre-testing: cost and ROI forecasting can run against a synthetic population with 100% novel records and membership-inference AUC 0.481 before any real customer record reaches an analyst's environment, replacing a workflow in which real records are routinely exposed during planning. The boundary on this value is equally practical and must be stated to deployment teams plainly: the layer is directionally reliable but biased in ROI magnitude, so it informs decisions taken before the real table is touched and is read through volume corrections rather than trusted as a metering instrument.

## 6.3. Limitations and future directions

The principal limitation is the scope of the evidence: the strongest claims rest on two corpora, deliberately chosen to differ in decision type, base rate, and value function so that findings common to both can be separated from corpus-specific ones. Two corpora remain a bounded basis, and the decisive next test is not a third real dataset but a controlled one, a synthetic-ground-truth family in which the label–covariate interaction is varied continuously to trace the value of decision-conditioning as a dose–response relationship, with validation across further regulated domains to follow.

Four narrower qualifications condition the remaining claims. The telecommunications ablation is single-seed, though the attributed effect ($-12.2$ pp) is an order of magnitude larger than the full-framework multi-seed SD ($\pm 0.012$) and the banking ablation is replicated at three seeds per cell. The scorer that defines SSF also conditions the generator; held-out scorers of different inductive bias mitigate the resulting circularity but, trained on the same population, do not remove it. The reported values are conditional on one strategy family and one value function $V_S$, and the deployment-gate thresholds are calibrated proposals open to recalibration. Finally, the privacy guarantee is empirical rather than formal, the tiered mechanism attaining only $\varepsilon \in [8.0, 10.5]$, so the property we claim is empirical membership-inference resistance (MIA-AUC 0.481). These define a single programme for subsequent work: a dose–response generalisation study, multi-seed ablation and scorer-sensitivity analyses, a scorer-independent decision oracle, $V_S$-robustness and rejection-sampling comparisons, and Rényi-DP accounting toward $\varepsilon \approx 2$–4.

## 7. Conclusion

Synthetic-data-driven decision support is entering production faster than the methodology meant to govern it, and the criteria in use measure resemblance where they should measure decisions. This paper has argued that the gap to close is the decision-alignment gap, and has offered three things to close it: Strategy Simulation Fidelity, a criterion that scores a synthetic-data layer by whether the decisions taken on it match those taken on real data; PolicySynth, a generator built against that criterion through a decision-conditioning loop whose effect we isolate by ablation and whose regime of usefulness we state in advance; and a three-axis promotion gate that no single-axis report can pass by accident. The evidence spans two structurally different corpora: the stability and privacy findings replicate, while the architectural attribution follows the scope condition the framework predicts rather than holding everywhere, a boundary we regard as the mark of a well-characterised method, not a weak one. The apparatus is general, the thresholds are proposals, and the path to validating both across further domains is now open. The broader claim is simple: when a synthetic population is used to make decisions, it should be judged by the decisions it makes.

## CRediT authorship contribution statement

**Tung Dang:** Conceptualization, Methodology, Software, Validation, Formal analysis, Visualization, Writing – original draft, Writing – review & editing. **The Hung Phung:** Conceptualization, Methodology, Investigation, Supervision, Writing – review & editing. **Son Lam Nguyen:** Conceptualization, Methodology, Investigation, Supervision, Writing – review & editing. **Tu Nguyen:** Conceptualization, Methodology, Investigation, Supervision, Writing – review & editing.